\documentclass[runningheads]{llncs}

\usepackage[T1]{fontenc}

\usepackage{hyperref}
\usepackage{graphicx}
\usepackage{subcaption}
\usepackage{amsmath}
\usepackage{amssymb}
\usepackage{booktabs}
\usepackage{wrapfig}
\usepackage{comment}
\usepackage{tikz}
\usetikzlibrary{patterns, positioning, calc, decorations.pathreplacing}
\usepackage{multirow}

\usepackage{color}

\begin{document}

\title{What Do Students Learn?\\
A Feature-Level Analysis of Dark Knowledge}

\titlerunning{What Do Students Learn?}

\author{Seungu Kang\orcidID{0009-0000-1003-0439} \and
Songkuk Kim\orcidID{0000-0003-4147-4627}}

\authorrunning{S. Kang and S. Kim}

\institute{Yonsei University, Seoul 03722, Republic of Korea\\
\email{\{ksw030721, songkukk\}@yonsei.ac.kr}}

\maketitle          

\begin{abstract}
Knowledge Distillation (KD) is a powerful tool for model compression, yet the precise mechanisms by which student models acquire feature representations remain underexplored. In this work, we analyze student feature learning using the Interaction Tensor framework. Our analysis reveals that effective KD acts as a regularizer that prunes low-frequency, sample-specific features, encouraging the student to rely on a compact set of highly reusable features. Crucially, we observe that the dataset-level confusion matrix contains structural information analogous to the teacher's "Dark Knowledge." Leveraging this insight, we propose Confusion Distillation (CD), a teacher-free self-distillation method that utilizes the model's own evolving confusion patterns as dynamic soft targets. CD achieves competitive performance on ResNet-34 and ResNet-50 for CIFAR-100, outperforming existing self-distillation methods like CS-KD and PS-KD by 1.2\% while offering a computationally efficient alternative to standard KD.

\end{abstract}

\section{Introduction}
As deep learning models have grown in scale, there have been continuous attempts to transfer knowledge learned in complex architectures to smaller models, enabling more efficient inference. Hinton \textit{et al.}~\cite{hinton2015distilling} proposed Knowledge Distillation (KD), which transfers the generalization structure of complex large-scale neural networks or ensemble models during training to smaller models, demonstrating that small models can also acquire generalization capabilities similar to those of large models. A key component of this process is dark knowledge, which refers to the similarity structure encoded in the teacher's predicted probabilities for the non-ground-truth classes. For instance, some incorrect classes may receive higher probabilities than others and such inter-class structure among incorrect labels reflects the classification boundaries the teacher has learned in the data space. In this way, KD serves not only as a model compression technique but also as a mechanism for transferring relational knowledge among classes, and has become established as a standard performance improvement method across various domains.

While recent studies have interpreted KD as a regularization mechanism that reduces prediction variance—similar to label smoothing~\cite{yuan2020revisiting,zhou2021rethinking}—these analyses primarily focus on the output distributions. Consequently, a concrete understanding of how student models construct representations in the feature space, and how this differs from independently trained models, remains underexplored. 

We analyze the feature-learning behavior of student models using the Interaction Tensor, a framework developed by Jiang \textit{et al}.~\cite{jiang2024interaction}. Unlike conventional analyses focused on output distributions, this framework captures the three-way interactions among models, data, and features. While the Interaction Tensor was originally proposed to explain the Generalization Disagreement Equality (GDE) phenomenon, we adopt this methodology to quantitatively visualize the structural changes KD induces in the student model’s feature representations.

Using this framework, we construct a single Interaction Tensor to compare 20 baseline models, 20 student models, and 20 teacher models on the same feature axis, and analyze how the three types of models utilize features across the dataset. Baseline models tend to learn a large number of low-frequency features that appear in only a small subset of data samples, whereas teacher models learn relatively fewer low-frequency features and exhibit higher activation frequencies for the learned features. Moreover, when classifying individual data points, teacher models rely on fewer features than baseline models and require a smaller number of features to achieve the same level of prediction confidence.

Our analysis reveals that student models closely mimic the teacher’s feature learning behavior. The teacher provides regularization that reduces feature variance, allowing students to achieve high confidence with fewer core features—a strategy isolated models struggle to develop. Hypothesizing that the dataset-level confusion structure captures meaningful inter-class relationships similar to a teacher's knowledge, we propose \textit{Confusion Distillation}. This self-distillation method utilizes the model's own confusion matrix, updated via exponential moving average, as a dynamic soft target. By converting confusion patterns into supervisory signals, our approach guides the model toward effective feature utilization without an external teacher. We compare this approach against other self-distillation methods, CS-KD~\cite{yun2020classwise} and PS-KD~\cite{kim2021self}. Our experiments show that confusion distillation outperforms these existing self-distillation techniques, introducing a new way to exploit confusion information directly as a knowledge distillation signal in multi-class classification. Our contributions are as follows:

\begin{itemize}
    \item To understand how knowledge distillation affects representation learning, we quantitatively analyze the feature usage structures of baseline, student, and teacher using the Interaction Tensor, and explained what changes KD induces at the feature level.
    
    \item Through this analysis, we show that small models have the potential to acquire teacher-like representation patterns when provided with appropriate supervisory signals, while confirming that such structures are difficult to obtain through standalone training.
    
    \item Building on these observations, we propose \emph{confusion distillation}, a self-distillation method that enables student models to adjust the direction of their representation learning without relying on an external teacher.
\end{itemize}

\section{Related Work}
\subsubsection{Knowledge Distillation} Since the seminal work of Hinton \textit{et al.}~\cite{hinton2015distilling}, Knowledge Distillation (KD) has become a fundamental technique for model compression. While the original KD transfers dark knowledge via soft logits, subsequent research has expanded this to feature-level distillation. FitNets~\cite{romero2015fitnets} introduced the concept of using intermediate feature maps as hints to guide the student. Following this, Attention Transfer (AT)~\cite{zagoruyko2017paying} proposed aligning the attention maps of the teacher and student, while Factor Transfer~\cite{kim2018paraphrasing} focused on transferring paraphrased, compact information distilled from the teacher's feature maps. More recently, Contrastive Representation Distillation (CRD)~\cite{tian2020contrastive} utilized contrastive learning to maximize the mutual information between the two networks, and Relational Knowledge Distillation (RKD)~\cite{park2019relational} emphasized transferring the structural relations of data examples rather than individual features.

\subsubsection{Analysis of Neural Representations} Understanding the internal representations of deep networks is crucial for interpreting KD. SVCCA~\cite{raghu2017svcca} and CKA~\cite{kornblith2019similarity} are widely used metrics to measure the similarity between layer representations of different neural networks. While these methods quantify similarity, they do not explicitly reveal the mechanism of feature utilization for individual samples. Jiang \textit{et al.}~\cite{jiang2024interaction} introduced the Interaction Tensor, a framework that decomposes the interaction between models, data, and features. We adopt this framework to provide a fine-grained analysis of the "Dark Knowledge" mechanism in KD.

\subsubsection{Self-Distillation} 
Self-distillation aims to improve a model's performance without a pre-trained teacher network. Early works like Born-Again Neural Networks (BAN)~\cite{furlanello2018born} trained students sequentially using the previous generation as a teacher. Deep Mutual Learning (DML)~\cite{zhang2018deep} trained multiple networks simultaneously by learning from each other. Be Your Own Teacher (BYOT)~\cite{zhang2019be} improved performance by distilling knowledge from deeper layers to shallower layers within a single network. Recently, regularization-based approaches have gained attention. Teacher-free KD (Tf-KD)~\cite{yuan2020revisiting} showed that self-training with soft labels acts as label smoothing regularization, while DDGSD~\cite{xu2019self} utilized data distortion to generate diverse targets. CS-KD~\cite{yun2020classwise} and PS-KD~\cite{kim2021self} utilized class-wise predictions and progressive targets, respectively. More recently, unified frameworks~\cite{yang2023from} have been proposed to bridge the gap between standard KD and self-KD through normalized loss functions. Our proposed Confusion Distillation can be viewed as a self-distillation approach that leverages class-level prediction distributions derived from the training process.

\subsubsection{Learning from Confusion}
While confusion matrices are primarily used for evaluation, several studies have incorporated them into the training objective. Early works~\cite{koco2013multi} proposed minimizing the norm of the confusion matrix for multi-class classification. In specific domains, confusion-aware architectures~\cite{Yan2019confusion} have been designed to reduce inter-class ambiguity. Recent advances in optimization have also sought to bridge the gap between evaluation metrics and loss functions by differentiating through confusion matrix-based metrics~\cite{tsoi2022bridging,Han2024anyloss}. However, these approaches typically use confusion information to optimize classification metrics directly or to design auxiliary losses. In contrast, our work reinterprets the confusion structure as a distinct source of "Dark Knowledge" and utilizes it as a soft distillation target to guide feature learning.

\section{Analyzing Knowledge Distillation}
\subsection{Experimental Settings}
We compare a ResNet-18 student and baseline against a ResNet-152 teacher~\cite{he2016resnet}. Experiments are conducted on CIFAR-100, which offers a balanced complexity for analyzing feature interactions compared to CIFAR-10 or ImageNet. Models are optimized using SGD with momentum and a weight decay of $5 \times 10^{-4}$. We employ a multi-step learning rate scheduler with linear warm-up~\cite{goyal2017accurate}. Standard data augmentations, including random cropping and horizontal flipping, are applied, following the standard practice for CIFAR datasets. The student model is trained with a temperature of 2 and a soft-hard loss ratio of $0.85:0.15$.

\subsection{Interaction Tensor}
Following the framework proposed by Jiang \textit{et al.}~\cite{jiang2024interaction}, we construct the Interaction Tensor to analyze feature learning behaviors. We extract features by projecting the penultimate layer outputs onto their top-50 principal components. These features are then grouped into common clusters based on cross-model correlations to handle random initialization. Finally, by thresholding feature activations, we construct a binary tensor $\Omega \in \{0,1\}^{M \times N \times T}$, where an entry $\Omega_{m,n,t}=1$ indicates that model $m$ utilizes feature cluster $t$ to classify data sample $n$.

\subsection{Frequency of Feature Occurrence}

\begin{figure}[htb]
\centering
\begin{subfigure}[b]{0.45\textwidth}
    \includegraphics[width=\textwidth]{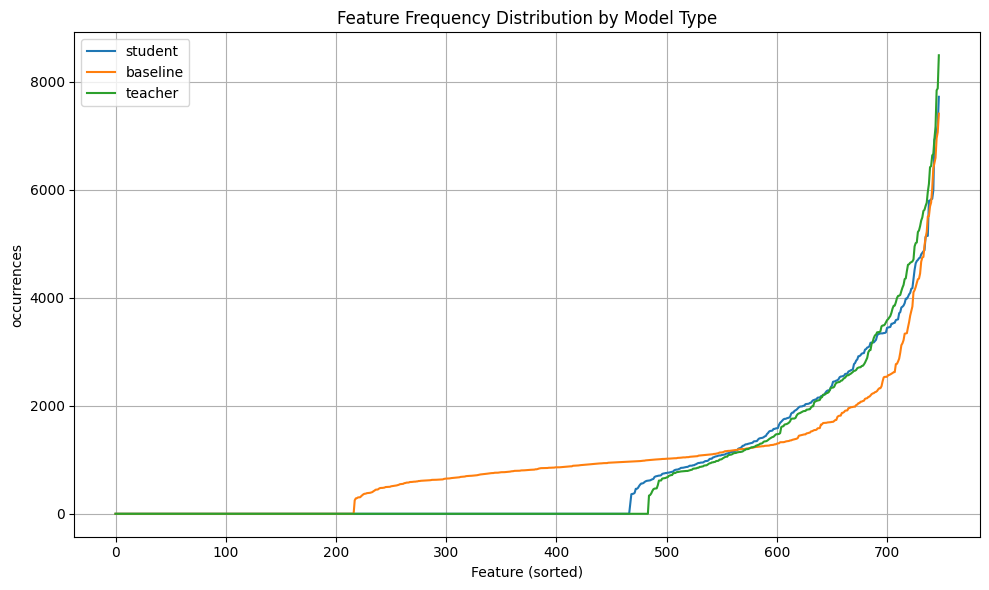} 
    \caption{}
    \label{fig:subfig1a}
\end{subfigure}
\hfill
\begin{subfigure}[b]{0.45\textwidth}
    \includegraphics[width=\textwidth]{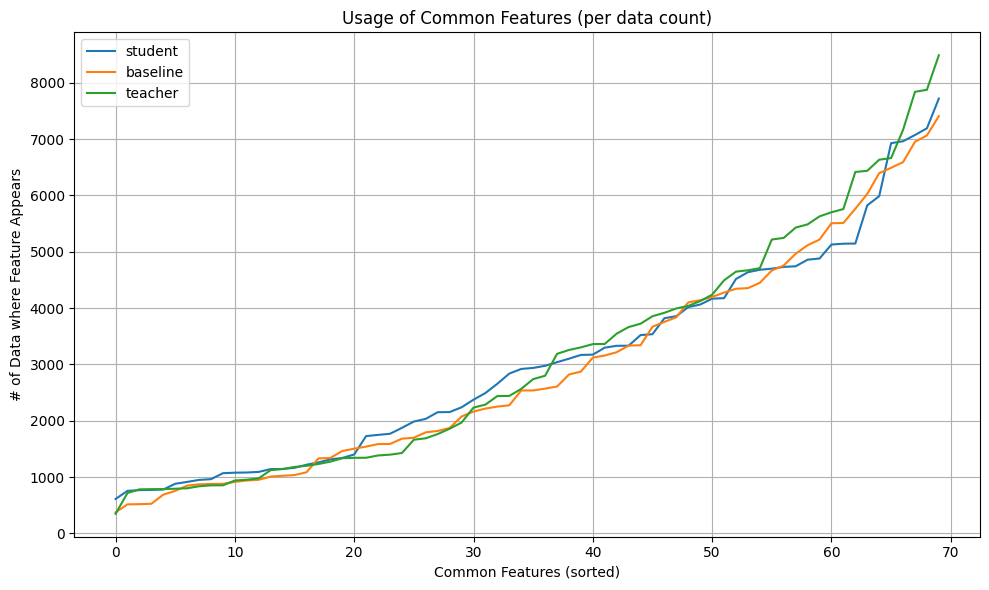} 
    \caption{}
    \label{fig:subfig1b}
\end{subfigure}

\caption{ \textbf{(a)} Feature frequency distributions for baseline, student, and teacher models. Features are sorted in ascending order based on the number of data points in which each feature appears. \textbf{(b)} Feature frequency distributions of commonly activated features shared by baseline, student, and teacher models }
\label{fig:fig1}
\end{figure}

To analyze which features were activated by which data points for each model (baseline, student, and teacher), we summed the Interaction Tensor along the model axis. The resulting data–feature matrix for each model was then binarized such that entries with nonzero values were set to 1, and all others were set to 0. Summing this binary matrix along the data axis yielded a feature frequency vector, representing the number of data points in which each feature was active. These frequency values were sorted in ascending order and visualized as shown in Figure~\ref{fig:subfig1a}.

The baseline model trained without distillation exhibits a distribution pattern consistent with the findings of Jiang \textit{et al.}~\cite{jiang2024interaction}. Most features appear very infrequently, while only a small number of features occur with high frequency, forming a long-tailed distribution. This pattern suggests that the model learns a large number of local, data-dependent features. This observation aligns with the theoretical finding that deep networks must memorize rare, long-tailed patterns to achieve good generalization performance when the data distribution is long-tailed~\cite{feldman2020does}.
Among them, high-frequency features that are repeatedly activated across many data points are more likely to contribute to the discriminative structure of the model, as they are representations that the model consistently refers to during classification; these features are selected via PCA, which captures the dominant variance directions in the data. In contrast, relying on a large number of low-frequency features for image classification is less compelling, as such features are more likely to correspond to sample-specific patterns and are therefore difficult to interpret as generalizable discriminative cues. 

The teacher model shows a substantial reduction in the number of low-frequency features and overall learns a smaller set of features. The activation frequencies of the learned features increase overall. Rather than indicating that the teacher learns more sample-specific features, this suggests that it acquires more generalized features that are reusable across a large number of data points. The feature frequency distribution of the student model closely resembles that of the teacher model; the two curves largely overlap across the entire range.

More concretely, the numbers of learned features for each model are $|F_{\mathrm{baseline}}|=531$, $|F_{\mathrm{student}}|=281$, and $|F_{\mathrm{teacher}}|=264$, showing that the student model learns $|F_{\mathrm{baseline}}|-|F_{\mathrm{student}}|=250$ fewer features than the baseline. Here, $F_{\mathrm{model}} = \{ F_{\mathrm{model}}^{(1)}, F_{\mathrm{model}}^{(2)}, \dots, F_{\mathrm{model}}^{(|F_{\mathrm{model}}|)} \}$ denotes the set of feature appearance frequencies for a given model, where each element $F_{\mathrm{model}}^{(i)}$ represents the number of data samples in which the $i$-th feature is activated. The average frequencies of feature activation are $E[|F_{\mathrm{baseline}}|] \approx 1329$ for the baseline model, $E[F_{\mathrm{student}}] \approx 2121$ for the student model, and $E[|F_{\mathrm{teacher}}|] \approx 2243$ for the teacher model. When restricting the analysis to the 70 features commonly activated by all three models, i.e., $|F_{\mathrm{baseline}} \cap F_{\mathrm{student}} \cap F_{\mathrm{teacher}}| = 70$, the average activation frequencies are $E[|F_{\mathrm{baseline|shared}}|] \approx 2938$, $E[|F_{\mathrm{student|shared}}|] \approx 3043$, and $E[|F_{\mathrm{teacher|shared}}|] \approx 3118$. As shown in Figure~\ref{fig:subfig1b}, the activation frequencies of these shared features are very similar across models, with the student and teacher models exhibiting slightly higher usage. This suggests that distillation not only encourages the learning of high-frequency features but also promotes more intensive reuse of the same features during inference.

In addition, the numbers of features shared between each pair of models are $|F_{\mathrm{baseline}} \cap F_{\mathrm{student}}| = 132$, $|F_{\mathrm{baseline}} \cap F_{\mathrm{teacher}}| = 96$, and $|F_{\mathrm{student}} \cap F_{\mathrm{teacher}}| = 170$. Despite differences in model architecture, the student and teacher models share a larger number of features. Performance improvements in KD are better explained by the student acquiring feature representations that closely align with those of the teacher model, rather than by learning entirely novel features. Such alignment is facilitated by the more specific guidance provided by the teacher’s soft targets during feature learning.

\subsection{Features and Confidence level}
\begin{figure}[htb]
\centering

\begin{subfigure}[b]{0.45\textwidth}
    \centering
    \includegraphics[width=\textwidth]{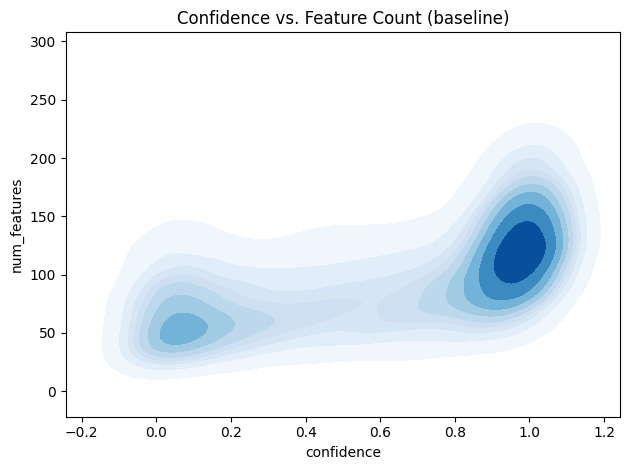}
    \caption{}
    \label{fig:subfig2a}
\end{subfigure}
\hfill
\begin{subfigure}[b]{0.45\textwidth}
    \centering
    \includegraphics[width=\textwidth]{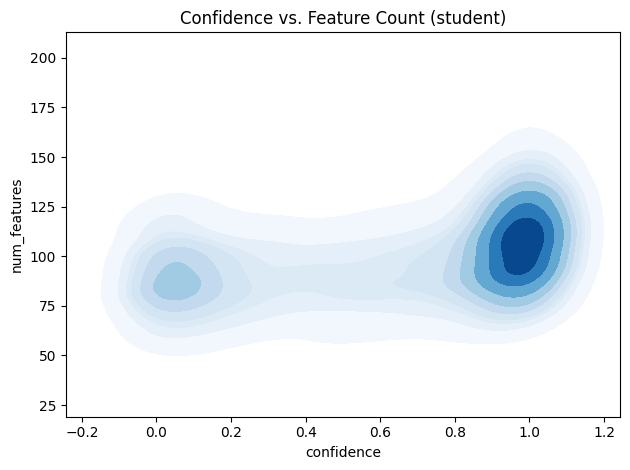}
    \caption{}
    \label{fig:subfig2b}
\end{subfigure}

\begin{subfigure}[b]{0.45\textwidth}
    \centering
    \includegraphics[width=\textwidth]{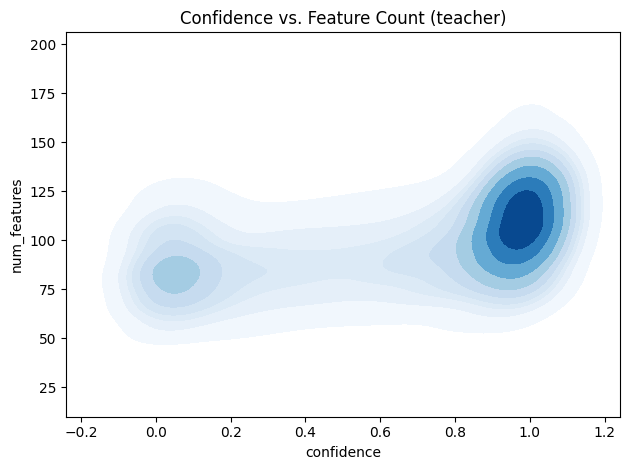}
    \caption{}
    \label{fig:subfig2c}
\end{subfigure}
\hfill
\begin{subfigure}[b]{0.45\textwidth}
    \centering
    \includegraphics[width=\textwidth]{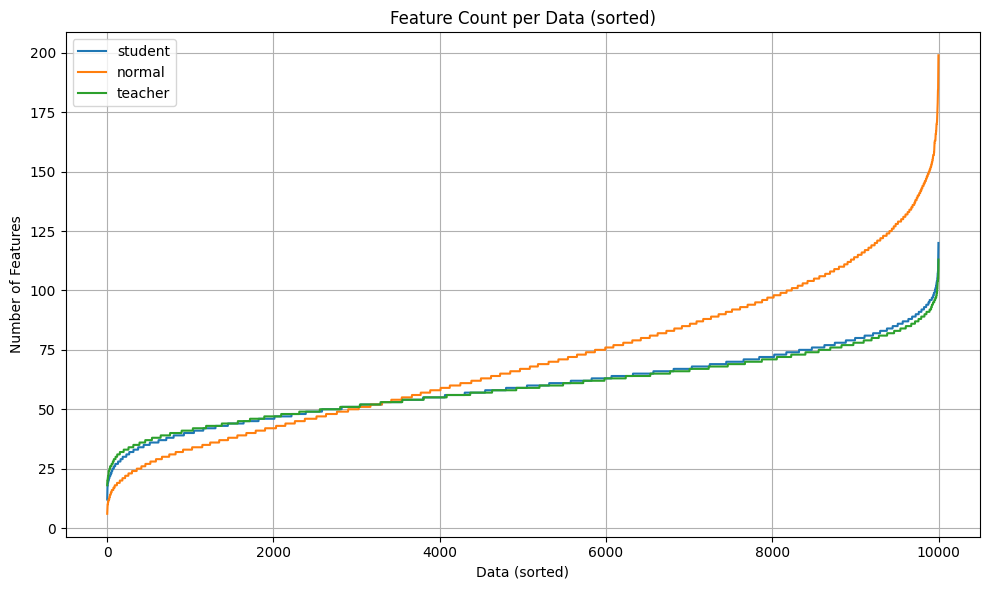}
    \caption{}
    \label{fig:subfig2d}
\end{subfigure}

\caption{ A 2D kernel density estimation plot showing the relationship between model confidence and the number of features used for each data point. \textbf{(a)} Baseline models, \textbf{(b)} student models, and \textbf{(c)} a teacher model. \textbf{(d)} Comparison of the number of activated features per data point for baseline, student, and teacher models. Data points are also sorted in ascending order based on the number of activated features}
\label{fig:fig2}
\end{figure}

To generate the KDE plots, we computed the confidence of each data point as the softmax probability assigned to the correct class. We then counted the number of activated features for each data point and used these two quantities as the axes of the 2D kernel density estimation plot.
To investigate the relationship between prediction confidence and feature usage, we visualized the joint distribution of the number of active features and the softmax probability of the correct class (Figure~\ref{fig:fig2}). Prior work by Jiang \textit{et al.}~\cite{jiang2024interaction} on CIFAR-10 observed a "sparsity-confidence" correlation, where high-confidence predictions relied on fewer active features. In contrast, our analysis on CIFAR-100 reveals an inverse trend: high-confidence samples consistently activate a larger number of features compared to low-confidence samples.

We attribute this discrepancy to the Evidence Accumulation hypothesis required for fine-grained classification. In coarser tasks like CIFAR-10, a single dominant feature (e.g., a wheel) may be sufficient to confidently classify a "Car." However, in the fine-grained CIFAR-100 setting, distinguishing semantically similar classes (e.g., "Beaver" vs. "Otter") requires a coalition of complementary features. Relying on a sparse feature set results in ambiguity and lower confidence, whereas high confidence is achieved only when multiple features converge to rule out competing classes.

Figure~\ref{fig:fig2} demonstrates that the feature-confidence distribution of the student model is closely aligned with that of the teacher, deviating from the baseline in two critical aspects. First, the student exhibits greater efficiency in high-confidence regimes: as shown in Figure~\ref{fig:subfig2b}, it maintains a notably lower upper bound on feature usage for high-confidence samples compared to the baseline. Second, for a reduced number of low-confidence (difficult) samples, the student actually increases its feature usage (raising the lower bound).

Synthesizing these observations, the student appears to adopt a dynamic strategy. For difficult cases, it mitigates uncertainty by combining more features to accumulate evidence. However, for the majority of data, it achieves evidential sufficiency with a significantly more compact representation. This indicates that distillation does not simply encourage feature memorization, but rather fosters the construction of potent feature coalitions—enabling the student to achieve high confidence using fewer, but more discriminative, features.

We compute the number of active features for each data point by summing the binarized data–feature matrix along the feature axis to quantify how many features were actually activated per sample. The resulting distributions are visualized in Figure~\ref{fig:subfig2d}, which shows that the student model uses fewer features than the baseline model when classifying an individual image. To further compare the two models, we calculate the average number of features used per data point by aggregating all feature activations and dividing by the number of samples used to construct the Interaction Tensor,

\begin{equation}
    \bar{D}_{\mathrm{model}}=\frac{\sum_i{F_{\mathrm{model}}^{(i)}}}{|\mathcal{D}|}.
\end{equation}

Based on this calculation, the baseline models use $\bar{D}_{\mathrm{baseline}} \approx 71$ features on average, whereas the student and teacher models use only $\bar{D}_{\mathrm{student}} \approx 60$ and $\bar{D}_{\mathrm{teacher}} \approx 59$.
This indicates that the student not only learns fewer features overall but also makes more effective use of the features it has learned.
 
Therefore, the high-frequency features learned by the student model are not merely patterns that appear often—such as background information—but features that contribute to image classification and suggest that the model has learned to combine a smaller set of features more effectively. In other words, the soft targets provided by the teacher model act as a form of feature-level regularization that reduces the variance of the features being learned. This mechanism leads the student model to suppress low-frequency, less informative features and instead acquire generalized, high-frequency features, ultimately guiding it to classify inputs using a compact yet effective combination of features. Just as a human teacher offers curriculum, the teacher model directs the student model to focus on learning useful features. This process explains why the student model, despite sharing the same architecture as the baseline model, achieves better performance.

\section{Confusion Distillation}
\subsection{Student's Confusion As Dark Knowledge}
Through Interaction Tensor analysis, we confirm that the student model learns fewer low-frequency features and adopts more effective combinations of features. This naturally leads to the next question: How can we provide appropriate guidance to a student model in the absence of a teacher? In Knowledge Distillation, the teacher model’s softmax output is typically used as the soft target that guides the training of the student model. The softmax distribution reflects inter-class similarity and represents the essential form of dark knowledge. Recent studies suggest that dark knowledge primarily acts as a strong regularization term~\cite{tang2020understanding} and that soft labels facilitate tighter clustering of same-class representations in the feature space~\cite{muller2019when}. Motivated by these observations, we hypothesize that the similarity structure among data classes represents teacher-independent properties inherent to the dataset itself and the confusion matrix aggregates prediction tendencies across the entire dataset, thereby capturing more generalized inter-class information. To examine this hypothesis, we evaluate whether the teacher model’s average class-wise softmax outputs resemble the confusion ratio produced by the baseline model.

Let $\hat{p}^{(n)} \in \mathbb{R}^K$ denote the teacher’s softmax output for a sample $n$ from the testset $\mathcal{D}$. 
For each class $j$, we compute the teacher’s class-wise average softmax matrix $M \in \mathbb{R}^{K \times K}$ over the subset 
$\mathcal{D}_j = \{ n \in \mathcal{D} \mid y_n = j \}$ as:
\begin{equation}
M_{j,i} = \frac{1}{|\mathcal{D}_j|} \sum_{n \in \mathcal{D}_j} \hat{p}^{(n)}_i, 
\quad j,i = 1,\dots,K.
\end{equation}
Each row $M_{j,:}$ thus represents the mean softmax probability distribution produced by the teacher when the true class is $j$.

The baseline’s confusion ratio matrix $Q \in \mathbb{R}^{K \times K}$ is then defined as
\begin{equation}
Q_{j,i}
=
\frac{1}{|\mathcal{D}_j|}
\sum_{n \in \mathcal{D}_j}
\mathbb{I}\!\left(\hat{y}_n^{(t)} = i\right),
\quad i,j = 1,\dots,K,
\end{equation}

where each row $Q_{j,:}$ represents the class-wise prediction distribution of the student model for samples with true class $j$. We quantitatively compare the structural similarity between the teacher’s average softmax matrix $M$ and the off-diagonal components of the student’s confusion ratio $Q$.
First, we measure the cosine similarity for each class after excluding the self-class component. 
The resulting cosine similarity has a mean value of approximately 0.76, indicating that while the two distributions are not identical in the high-dimensional space, they largely share similar directions. 
This suggests that the confusion ratio does not precisely reproduce the fine-grained probability values of the teacher’s softmax distribution, but instead captures a coarse inter-class similarity structure. 
To assess global correlation, we compute the Pearson correlation coefficient, which yielded a value of approximately 0.85 under the self-class-excluded setting. 
This indicates that the two matrices share a strong overall trend in their inter-class confusion patterns.

Further quantifying the degree of structural overlap between the teacher’s class-wise average softmax distribution and the baseline model’s confusion ratio, we measure the \textit{Jaccard similarity index}. 
Focusing exclusively on inter-class relationships, the self-class component was excluded from the comparison, and the Jaccard similarity for each class $j$ was computed as
\begin{equation}
J_j
=
\frac{\sum_{i \neq j} \min\left(M_{j,i}, {Q}_{j,i}\right)}
     {\sum_{i \neq j} \max\left(M_{j,i}, {Q}_{j,i}\right)}.
\end{equation}

The overall degree of structural overlap was then evaluated by averaging the Jaccard similarity across all classes:
\begin{equation}
\bar{J}
=
\frac{1}{K} \sum_{j=1}^{K} J_j.
\end{equation}

The mean Jaccard similarity under the self-class-excluded setting was observed to be approximately 0.38. 
Although the teacher’s softmax distribution and the baseline confusion ratio differ at the level of individual probability values, this result indicates that the allocation of probability mass across classes is not random but exhibits partial structural overlap. Therefore, it serves not merely as a statistical record of misclassifications but as a meaningful soft target analogous to the dark knowledge traditionally provided by the teacher model in KD.

\subsection{Method}

In this section, we propose a training framework in which the model is first trained with a standard Cross-Entropy (CE) loss during the initial phase, and then gradually switched to a self-distillation stage that utilizes soft labels derived from the confusion matrix.

\subsubsection{Early Training: Cross-Entropy Loss}

At the early stage of training, the predicted probability distribution is not yet stable, and directly using the confusion matrix as a soft label can introduce significant noise. 
Therefore, during the initial epochs, we employ the stadard cross-entropy loss ($L_{\mathrm{CE}}$).

\subsubsection{Confusion Ratio Update}

Since the confusion ratios obtained immediately after the transition epoch are noisy and unstable, 
we employ two mechanisms to stabilize the confusion-based soft targets. First, the confusion ratio is initialized with a smoothing matrix $S$, which acts as a prior to prevent unreliable early estimates from dominating training. Second, instead of directly using the raw confusion ratio $Q^{(t)}$ at each epoch, we maintain an Exponential Moving Average (EMA) estimate $\tilde{Q}^{(t)}$, which aggregates historical information and mitigates fluctuations arising from unstable prediction statistics. This EMA strategy, inspired by the Mean Teacher framework~\cite{tarvainen2017mean}, ensures smoother and more stable distillation. The update rule is defined as follows:

\begin{equation}
\tilde{Q}^{(t_{\text{switch}}-1)} = S, \quad
\tilde{Q}^{(t)} = \mu \tilde{Q}^{(t-1)} + (1-\mu) Q^{(t)}, \quad t \geq t_{\text{switch}}.
\end{equation}

The smoothing matrix $S \in \mathbb{R}^{K \times K}$ is defined as:

\begin{equation}
S_{ij} =
\begin{cases}
1 - \epsilon + \dfrac{\epsilon}{C}, & \text{if } i = j, \\[6pt]
\dfrac{\epsilon}{C}, & \text{if } i \neq j,
\end{cases}
\end{equation}

where $\mu = 0.9$ is the momentum coefficient. 

\subsubsection{Loss Function}

Each row $\tilde{Q}^{(t)}_{k,:}$ of the EMA-smoothed confusion ratio $\tilde{Q}^{(t)}$ is used as a soft label distribution for class $k$. 
The soft target loss is defined as:

\begin{equation}
L_{\text{soft}}^{(t+1)} = KL(\tilde{Q}_{k,\cdot}^{(t)} \parallel p_T) \cdot T^2,
\end{equation}

where $p_T$ denotes the student model’s softened prediction obtained by applying temperature scaling to its output logits $\mathbf{z}$. After the transition phase, the total loss function combines the confusion-based soft label loss and the conventional hard label loss:

\begin{equation}
\mathcal{L} = \alpha \cdot L_{\text{soft}} + \beta \cdot L_{\text{CE}},
\end{equation}

where $\alpha$ and $\beta$ are weighting coefficients that balance the contributions of the soft label loss and the hard label loss, respectively.

\subsection{Effects of Confusion Distillation}
\begin{figure}[htb]
\centering
\begin{subfigure}[b]{0.45\textwidth}
    \includegraphics[width=\textwidth]{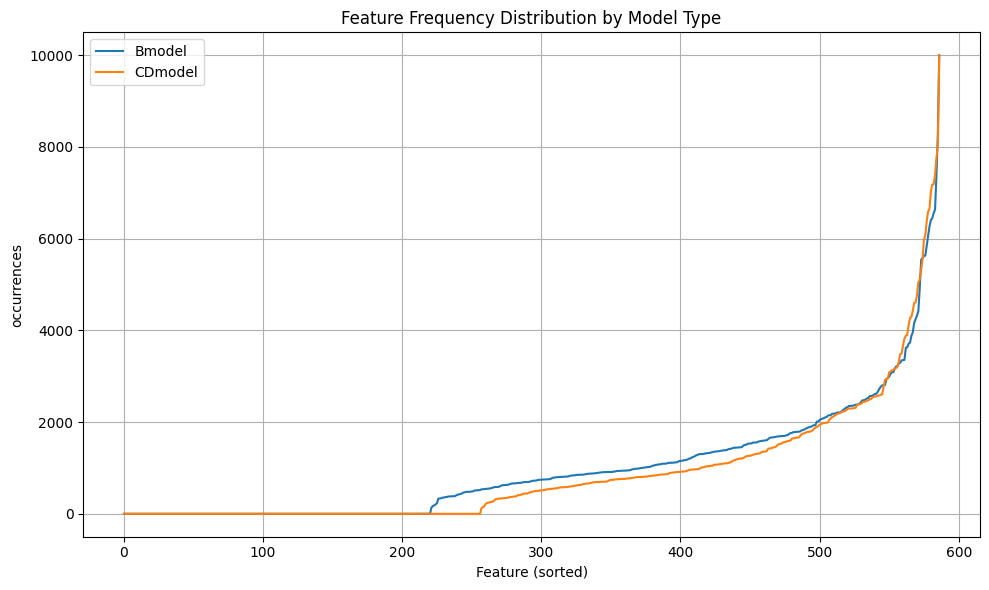} 
    \caption{}
    \label{fig:subfig3a}
\end{subfigure}
\hfill
\begin{subfigure}[b]{0.45\textwidth}
    \includegraphics[width=\textwidth]{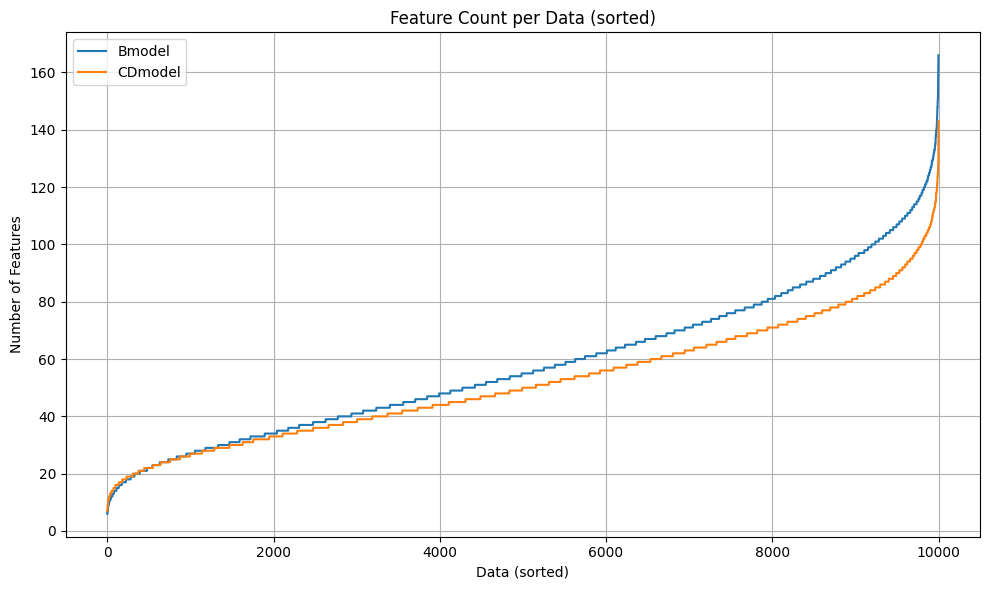} 
    \caption{}
    \label{fig:subfig3b}
\end{subfigure}

\caption{ \textbf{(a)} Feature frequency distributions for baseline, CD models. \textbf{(b)} Comparison of the number of activated features per data point for baseline, CD models.}
\label{fig:fig3}
\end{figure}

To examine whether Confusion Distillation (CD) produces effects similar to those of conventional Knowledge Distillation (KD), we analyze how the CD-trained model learns and utilizes features using the Interaction Tensor. Following the same procedure used in the KD analysis, we first measure how frequently each feature was activated across the dataset. The CD model showed a reduction in the total number of active features, decreasing from $\#F_B = 336$ to $\#F_{CD} = 330$, and it suppressed many of the low-frequency features that the baseline model continued to rely on.

Next, we examine how many features were used for each individual data point. As shown in Figure~\ref{fig:subfig3a}, the CD model uses fewer features than the baseline model when classifying a single image. However, the features it does use tend to be activated more frequently across the dataset, and the average activation count is higher for the CD model ($E[F_B] \approx 2724$ vs. $E[F_{CD}] \approx 3021$). Even when we restrict the comparison to features shared by both models, the CD model still shows higher activation frequencies ($E[F_{baseline|\text{shared}}] \approx 2183$, $E[F_{CD|\text{shared}}] \approx 2389$), indicating that CD reuses a smaller feature set more efficiently.

These patterns align with the effects observed under KD and suggest that CD induces a similar phenomenon to the feature-level regularization seen in KD-trained student models. Although CD does not rely on the teacher’s softmax distribution, it appears to convey a form of dark knowledge analogous to that of KD. Through the structural information embedded in the confusion ratio, the CD model learns for itself which features are useful and which should be suppressed.

\subsection{Hyperparameter Tuning}
\begin{table}[t]
    \centering
    \begin{minipage}{0.65\columnwidth}
    \centering
    \resizebox{\columnwidth}{!}{
    \begin{tabular}{lccccc}
    \toprule
    \textbf{Model} & \textbf{Soft:Hard} & \textbf{Epoch} & \textbf{Transition Schedule} & \textbf{Top-1} & \textbf{Top-5} \\
    \midrule
    \multirow{11}{*}{ResNet-18} & 0:1 & 200 & -- & 75.61 & 93.05 \\
             & 0.85:0.15 & 200 & 3 : \textcolor{blue}{17} & 66.34 & 89.07 \\
             & 0.85:0.15 & 200 & 6 : \textcolor{blue}{14} & 73.47 & 91.75 \\
             & 0.5:0.5 & 200 & 3 : \textcolor{blue}{3} : 3 : \textcolor{blue}{3} : 2 : \textcolor{blue}{2} : 2 : \textcolor{blue}{2} & 76.11 & 93.17 \\
             & 0.85:0.15 & 200 & 3 : \textcolor{blue}{3} : 3 : \textcolor{blue}{3} : 2 : \textcolor{blue}{2} : 2 : \textcolor{blue}{2} & 75.64 & 92.23 \\
             & 0.7:0.3 & 200 & 3 : \textcolor{blue}{3} : 3 : \textcolor{blue}{3} : 2 : \textcolor{blue}{2} : 2 : \textcolor{blue}{2} & 76.44 & 92.54 \\
             & 0.7:0.3 & 200 & 3 : \textcolor{blue}{3} : 3 : \textcolor{blue}{3} : 8 & 76.96 & 93.91 \\ & 0.7:0.3 & 200 & 14 : \textcolor{blue}{2} : 2 : \textcolor{blue}{2} & 75.78 & 92.62 \\
             & 0.7:0.3 & 200 & 2 : \textcolor{blue}{2} : 4 : \textcolor{blue}{2} : 10 & 76.75 & 93.56 \\
             & 0.7:0.3 & 300 & 3 : \textcolor{blue}{3} : 6 : \textcolor{blue}{3} : 15 & \textbf{77.14} & 93.53 \\
             & 0.7:0.3 & 300 & 4.5 : \textcolor{blue}{4.5} : 4.5 : \textcolor{blue}{4.5} : 12 & 76.93 & 94.02 \\
    \midrule
    \multirow{2}{*}{ResNet-34} & 0:1 & 200 & -- & 76.76 & 93.37 \\
             & 0.7:0.3 & 300 & 3 : \textcolor{blue}{3} : 6 : \textcolor{blue}{3} : 15 & 78.84 & 94.65 \\
    \midrule
    \multirow{2}{*}{ResNet-50} & 0:1 & 200 & -- & 77.39 & 93.96 \\
             & 0.7:0.3 & 300 & 3 : \textcolor{blue}{3} : 6 : \textcolor{blue}{3} : 15 & 79.70 & 95.00 \\
    \bottomrule
    \end{tabular}}
    \label{tab:hyperparameter}
    \end{minipage}
    \caption{Hyperparameter tuning results on CIFAR-100. Black and \textcolor{blue}{blue} denote hard-target and confusion-target learning phase respectively.}
\end{table}

To determine the optimal configuration of CD, 
We explore various \textit{soft–hard loss ratios} and \textit{transition schedules} to identify the most effective setting. The \textit{transition schedule} is represented as a ratio that defines the relative duration of each training stage, while the absolute length of each stage is determined by the total number of training epochs.

The best results were obtained when the \textit{soft–hard loss ratio} was set to $0.7 : 0.3$ and the \textit{transition ratio} to \texttt{3:3:6:3:15} with 300 training epochs.
\begin{tikzpicture}[font=\sffamily\small]

\definecolor{softblue}{RGB}{70, 130, 180} 

\def\h{1.2} 

\def\scale{0.35}

\coordinate (S0) at (0,0);
\coordinate (S1) at ({3*\scale},0);   
\coordinate (S2) at ({6*\scale},0);   
\coordinate (S3) at ({12*\scale},0);  
\coordinate (S4) at ({15*\scale},0);  
\coordinate (S5) at ({30*\scale},0);  


\fill[gray!20] (S0) rectangle ++({3*\scale},\h);
\draw[thick, black] (S0) rectangle ++({3*\scale},\h);
\node[black] at ($ (S0)!0.5!(S1) + (0, \h/2) $) {Hard};

\fill[softblue!20] (S1) rectangle ++({3*\scale},\h);
\draw[thick, softblue] (S1) rectangle ++({3*\scale},\h);
\node[softblue] at ($ (S1)!0.5!(S2) + (0, \h/2) $) {Soft};

\fill[gray!20] (S2) rectangle ++({6*\scale},\h);
\draw[thick, black] (S2) rectangle ++({6*\scale},\h);
\node[black] at ($ (S2)!0.5!(S3) + (0, \h/2) $) {Hard};

\fill[softblue!20] (S3) rectangle ++({3*\scale},\h);
\draw[thick, softblue] (S3) rectangle ++({3*\scale},\h);
\node[softblue] at ($ (S3)!0.5!(S4) + (0, \h/2) $) {Soft};

\fill[gray!20] (S4) rectangle ++({15*\scale},\h);
\draw[thick, black] (S4) rectangle ++({15*\scale},\h);
\node[black] at ($ (S4)!0.5!(S5) + (0, \h/2) $) {Hard Target (Stable)};

\draw[->, thick] (0,0) -- ({30*\scale + 0.5},0) node[right] {Epochs};

\foreach \x/\label in {0/0, 3/30, 6/60, 12/120, 15/150, 30/300} {
    \draw ({ \x*\scale }, 0) -- ({ \x*\scale }, -0.2);
    \node[below] at ({ \x*\scale }, -0.2) {\footnotesize \label};
}

\node[anchor=south west, align=left] at (0, \h+0.5) {
    \textbf{Phase Alternation Schedule (3:3:6:3:15)} \\
    \footnotesize \textcolor{black}{\textbf{Hard}}: CE Loss Only ($\mathcal{L}_{CE}$),
    \footnotesize \textcolor{softblue}{\textbf{Soft}}: Confusion Distillation + CE ($\alpha\mathcal{L}_{soft} + \beta\mathcal{L}_{CE}$)
};

\end{tikzpicture}

This configuration consistently improved both Top-1 and Top-5 accuracy across different ResNet architectures.
We also find that extending the confusion-based phase for too long caused accuracy degradation, implying that some confusion signals (e.g., bird$\xrightarrow{}$car) negatively affect discriminative learning and become increasingly reinforced as training progresses. 
Therefore, it is essential to carefully balance the transition schedule—allowing the model to learn from meaningful confusion information while avoiding overfitting to incorrect signals. 
After each confusion phase, reverting to hard-target training effectively re-aligns the model and stabilizes the learning process.

\subsection{Comparison with Other Self-Distillation Methods}
To validate the effectiveness of the proposed \textit{Confusion Distillation (CD)}, 
we compare it against existing self-knowledge distillation methods, 
\textit{CS-KD}~\cite{yun2020classwise} and 
\textit{PS-KD}~\cite{kim2021self}. 
Table~\ref{tab:comparison} presents the experimental results on CIFAR-100 with ResNet-18, ResNet-34, and ResNet-50 architectures.

Compared to the baseline and CS-KD~\cite{yun2020classwise}, the proposed CD consistently achieved improvements in both Top-1 and Top-5 accuracy. 
This suggests that CD does not simply mimic a teacher’s predictions but rather enables the model to recognize and exploit its own predictive uncertainty, 
transforming it into structural diversity within the feature space. 
While PS-KD~\cite{kim2021self} attained slightly higher performance with extended training (300 epochs), 
CD achieved comparable results with fewer epochs (200 epochs), indicating superior training efficiency. 
This demonstrates that confusion information can effectively complement representational learning even without explicit teacher supervision.

\begin{table}[t]
\centering
\resizebox{\columnwidth}{!}{
\begin{tabular}{lc cc cc cc cc}
\toprule
\multirow{2}{*}{\textbf{Method}} & \multirow{2}{*}{\textbf{Epoch}} 
  & \multicolumn{2}{c}{\textbf{ResNet-18}} 
  & \multicolumn{2}{c}{\textbf{ResNet-34}} 
  & \multicolumn{2}{c}{\textbf{ResNet-50}} 
  & \multicolumn{2}{c}{\textbf{DenseNet-121}} \\
\cmidrule(lr){3-4}\cmidrule(lr){5-6}\cmidrule(lr){7-8}\cmidrule(lr){9-10}
 & & Top-1 & Top-5 & Top-1 & Top-5 & Top-1 & Top-5 & Top-1 & Top-5 \\
\midrule
Baseline  & 200 & 75.86\tiny{±.09} & 92.90\tiny{±.23} & 77.61\tiny{±.35} & 93.76\tiny{±.04} & 78.48\tiny{±.56} & 94.53\tiny{±.36} & 79.03\tiny{±.11} & 94.78\tiny{±.09} \\
CS-KD     & 200 & 76.38\tiny{±.17} & 93.73\tiny{±.20} & 76.73\tiny{±.06} & 93.32\tiny{±.05} & 76.31\tiny{±.36} & 92.64\tiny{±.16} & 76.53\tiny{±.58} & 91.75\tiny{±.49} \\
PS-KD     & 300 & 77.41\tiny{±.22} & 94.16\tiny{±.15} & 77.33\tiny{±.12} & 94.34\tiny{±.05} & 78.41\tiny{±.31} & 94.93\tiny{±.25} & 79.84\tiny{±.23} & 95.40\tiny{±.09} \\
CD (Ours) & 200 & 76.85\tiny{±.10} & 93.82\tiny{±.09} & 77.87\tiny{±.08} & 94.21\tiny{±.07} & 78.63\tiny{±.21} & 94.78\tiny{±.09} & 78.71\tiny{±.18} & 94.80\tiny{±.08} \\
CD (Ours) & 300 & 77.13\tiny{±.01} & 93.37\tiny{±.12} & \textbf{78.53}\tiny{±.22} & \textbf{94.44}\tiny{±.17} & \textbf{79.38}\tiny{±.23} & \textbf{94.93}\tiny{±.07} & 79.64 \tiny{±.22} & 94.92\tiny{±.08} \\
\bottomrule
\end{tabular}}
\caption{Comparison of Confusion Distillation (CD) with other self-distillation methods on CIFAR-100.}
\label{tab:comparison}
\end{table}

Beyond numerical performance gains, CD provides empirical evidence that the confusion information can serve as a meaningful learning signal. 
Therefore, CD should not be viewed merely as another performance-boosting distillation method, but rather as an exploration of how confusion itself can act as a valuable signal that enriches representation learning 
and promotes generalization in teacher-free self-distillation frameworks.

\section{Conclusion}
This study used the Interaction Tensor to examine how knowledge distillation  changes the representation-learning behavior of student models. We show that student models suppress low-frequency features more than baseline models and rely on a smaller set of high-quality features for classification. This leads to a more compact and confident representation structure, reflecting a feature-level regularization effect induced by the teacher’s soft targets.

The Interaction Tensor analysis also indicates that small models rarely converge to such efficient feature structures on their own but can approach teacher-like representations when given appropriate supervisory signals. The strong similarity between the baseline model’s confusion patterns and the teacher’s class-wise average softmax outputs further suggests that confusion information itself contains meaningful inter-class relationships that can serve as an alternative supervisory cue.

Based on this insight, we introduce Confusion Distillation, which transforms confusion information from previous epochs into soft targets and mixes them with hard targets. Experiments across multiple ResNet architectures show that CD consistently improves performance and produces KD-like representational effects without a teacher.

For future work, the proposed approach may be extended to larger models or language models, enabling feature-level analyses across diverse architectures. Such extensions could further illuminate general principles of representation learning from the perspective of distillation.

\section{Acknowledgment}
This work was supported in part by Institute of Information \& communications Technology Planning \& Evaluation (IITP) grants (No. RS-2024-00395824, No.RS-2025-02214652) funded by the Korea Government (MSIT).

\end{document}